\documentclass[11pt]{article}
\usepackage{nodalida2019}
\usepackage{times}
\usepackage{url}
\usepackage{latexsym}
\usepackage[english]{babel}
\usepackage[utf8]{inputenc}
\usepackage[T1]{fontenc}

\usepackage{enumitem}
\setlist{nolistsep}
\aclfinalcopy % Uncomment this line for the final submission

\title{Nefnir: A high accuracy lemmatizer for Icelandic}

\author{Svanhvít Ingólfsdóttir, Hrafn Loftsson \\
  Department of Computer Science \\ \\
 Reykjavik University \\
  {\tt \string{svanhviti16, hrafn\string}@ru.is} \\\And
  Jón Daðason, Kristín Bjarnadóttir \\
  The Árni Magnússon Institute \\ for Icelandic Studies \\
  University of Iceland \\
  {\tt \string{jfd1, kristinb\string}@hi.is} \\}
  
\date{}

\begin{document}
\maketitle
\begin{abstract}
 Lemmatization, finding the basic morphological form of a word in a corpus, is an important step in many natural language processing tasks when working with morphologically rich languages. 
 We describe and evaluate \emph{Nefnir}, a new open source lemmatizer for Icelandic. Nefnir uses suffix substitution rules, derived from a large morphological database, to lemmatize tagged text.  Evaluation shows that for correctly tagged text, Nefnir obtains an accuracy of 99.55\%, and for text tagged with a PoS tagger, the accuracy obtained is 96.88\%. 
\end{abstract}

\section{Introduction}
%Here we should introduce the task of lemmatization and why it is useful.  If spae allows, we should also show an example of lemmatization for a given text (for Icelandic).

%Then we introduce Nefnir and mention what is special about it.

In text mining and Natural Language Processing (NLP), a \textit{lemmatizer} is a tool used to determine the basic form of a word (\textit{lemma}).  Lemmatization differs from \textit{stemming} in the way this base form is determined. While stemmers chop off word endings to reach the common stem of words, lemmatizers take into account the morphology of the words in order to produce the common morphological base form, i.e., the form of the word found in a dictionary.
This type of text normalization is an important step in pre-processing morphologically complex languages, like Icelandic, before conducting various tasks, such as machine translation, text mining and information retrieval.

%má stytta
To give an example from the Icelandic language, lemmatization helps find all instances of the personal pronoun \textit{ég} ``I'' in a text corpus, taking into account all inflectional forms (\textit{ég}, \textit{mig}, \textit{mér}, \textit{mín}, \textit{við}, \textit{okkur}, and \textit{okkar}). These variations of each word can be up to 16 for nouns and over a hundred for adjectives and verbs. The value of being able to reduce the number of different surface forms that appear for each word is therefore evident, as otherwise it is hard or even impossible to correctly determine word frequency in a corpus, or to look up all instances of a particular term. 

% meira motivation? 
In this paper, we describe and evaluate \textit{Nefnir} \cite{Nefnir}, a new open source lemmatizer for Icelandic. Nefnir uses suffix substitution rules derived (learned) from the Database of Modern Icelandic Inflection (DMII) \cite{Bjarnadottir_2012}, which contains over 5.8 million inflectional forms.

This new lemmatizer was used for large-scale lemmatization of the \textit{Icelandic Gigaword Corpus} \cite{Steingrimsson_2018} with promising results, but a formal evaluation had not been carried out. Our evaluation of Nefnir indicates that, compared to previously published results, it obtains the highest lemmatization accuracy of Icelandic, with 99.55\% accuracy given correct part-of-speech (PoS) tags, and 96.88\% accuracy given text tagged with a PoS tagger.
%SLI: Our evaluation of Nefnir shows that it reaches 99.55\% accuracy given correct part-of-speech (PoS) tags, and 96.88\% accuracy given text tagged with a PoS tagger, which indicates that it is the best available tool for lemmatization of Icelandic.

\section{Related work}
%Here we should discuss the common approaches to lemmatization in the literature.
%And have subsections for the lemmatizer that we know of for Icelandic.
The most basic approach to lemmatization is a simple look-up in a lexicon. This method has the obvious drawback that words that are not in the lexicon cannot be processed. To solve this, word transformation rules have been used to analyze the surface form of the word (the token) in order to produce the base form. These rules can either be hand-crafted or learned automatically using machine learning. %Both methods have been used extensively when creating lemmatizers. 

When hand-crafting the rules that are used to determine the lemmas, a thorough knowledge of the morphological features of the language is needed. This is a time-consuming task, further complicated in Icelandic by the extensive inflectional system \cite{Bjarnadottir_2012}. An example of a hand-crafted lemmatizer is the morphological analyzer that is part of the Czech Dependency Treebank \cite{Hajic_2018}.

Machine learning methods emerged to make the rule-learning process more effective, and various algorithms have been developed. These methods rely on training data, which can be a corpus of words and their lemmas or a large morphological lexicon \cite{Jongejan_2009}. By analyzing the training data, transformation rules are formed, which can subsequently be used to find lemmas in new texts, given the word forms. 

In addition, maching learning lemmatizers based on deep neural networks (DNNs) have recently emerged (see for example \textit{finnlem} \cite{Finnlem} for Finnish and \textit{LemmaTag} \cite{Kondratyuk_2008} for German, Czech and Arabic). Along with the best rule-derived machine learning methods, these are now the state-of-the-art approaches to lemmatizers for morphologically complex languages. % of stór fullyrðing? Vísa í heimild?

The biggest problem in lemmatization is the issue of unknown words, i.e. words not found in the training corpus or the underlying lexicon of the lemmatizer. This has been handled in various ways, such as by only looking at the suffix of a word to determine the lemma, thereby lemmatizing unseen words that (hopefully) share the same morphological rules as a known word \cite{Dalianis_2006}. DNN-based lemmatizers may prove useful in solving this issue, as they have their own inherent ways of handling these out-of-vocabulary (OOV) words, such as by using character-level context \cite{Bergmanis_2018}.

Previous to Nefnir, two lemmatization tools had been developed for Icelandic. We will now briefly mention these lemmatizers, before describing Nefnir further.

\subsection{CST Lemmatizer}
The CST Lemmatizer \cite{Jongejan_2009} is a rule-based lemmatizer that has been trained for Icelandic on the Icelandic Frequency Dictionary (IFD) corpus, consisting of about 590,000 tokens \cite{Pind_1991}. This is a language-independent lemmatizer that only looks at the suffix of the word as a way of lemmatizing OOV words, and can be used on both tagged and untagged input.

The authors of Lemmald (see Section \ref{lemmald}) trained and evaluated the CST Lemmatizer on the IFD and observed a 98.99\% accuracy on correctly tagged text and 93.15\% accuracy on untagged text, in a 10-fold cross-validation, where each test set contained about 60,000 tokens. Another evaluation of this lemmatizer for Icelandic \cite{Cassata_2007} reports around 90\% accuracy on a random sample of 600 words from the IFD, when the input has been PoS tagged automatically (with a tagging accuracy of 91.5\%).  The PoS tagger used was \emph{IceTagger} \cite{Loftsson_2008}, which is part of the IceNLP natural language processing toolkit \cite{Loftsson_2007}. These results indicate that the accuracy of this lemmatizer is very dependent upon the tags it is given. 
To our knowledge, the Icelandic CST Lemmatizer model is not openly available. 

\subsection{Lemmald} \label{lemmald}
The second tool is Lemmald \cite{Ingason_2008}, which is part of the IceNLP toolkit. It uses a mixed method of data-driven machine learning (using the IFD as a training corpus) and linguistic rules, as well as providing the option of looking up word forms in the DMII. Given correct PoS tagging of the input, Lemmald's accuracy measures at 98.54\%, in a 10-fold cross-validation. 
The authors note that the CST Lemmatizer performs better than Lemmald when trained on the same data, without the added DMII lookup. The DMII lookup for Lemmald delivers a statistically significant improvement on the accuracy (99.55\%), but it is not provided with the IceNLP distribution, so this enhancement is not available for public use.
%SLI: þetta kemur fram í greininni, samt engar nákvæmari tölur. 90% talan er bara fyrir MÍM
%A preliminary test indicates that when using IceTagger tags, the accuracy of Lemmald drops by around 1.5\% \cite{Ingason_2008}.
When used for lemmatization of the Icelandic Tagged Corpus (MÍM) \cite{Helgadottir_2012}, the lemmatization accuracy of Lemmald was roughly estimated at around 90\%.\footnote{See \url{https://www.malfong.is/index.php?lang=en&pg=mim}} 

\section{System Description}
The main difference between Nefnir and the two previously described  lemmatizers for Icelandic, CST Lemmatizer and Lemmald, is that Nefnir derives its rules from a morphological database, the DMII, whereas the other two are trained on a corpus, the IFD. Note that the IFD only consists of about 590,000 tokens, while the DMII contains over 5.8 million inflectional forms.

Nefnir uses suffix substitution rules, derived from the DMII to lemmatize tagged text. An example of such a rule is (\textit{ngar}, nkfn, \textit{ar}$\to$\textit{ur}), which can be applied to any word form with the suffix \textit{ngar} that has the PoS tag nkfn (a masculine plural noun in the nominative case), transforming the suffix from \textit{ar} to \textit{ur}. This rule could, for example, be applied to the word form \emph{kettlingar} ``kittens'' to obtain the corresponding lemma, \emph{kettlingur}. Words are lemmatized using the rule with the longest shared suffix and the same tag.

Each inflectional form in the DMII is annotated with a grammatical tag and lemma. As the DMII is limited to inflected words, the training data is supplemented with a hand-curated list of approximately 4,500 uninflected words (such as adverbs, conjunctions and prepositions) and abbreviations. 

To account for subtle differences between the tagsets used in the DMII and by the Icelandic PoS taggers, Nefnir translates all tags to an intermediate tagset which is a subset of both.

Rules are successively generated and applied to the training set, with each new rule minimizing the number of remaining errors. Rules continue to be generated until the number of errors cannot be reduced. The process is as follows:

\begin{enumerate}
\item Initially, assume that each word form is identical to its lemma.
\item Generate a list of rules for all remaining errors.
\item Choose the rule which minimizes the number of remaining errors and apply it to the training set, or stop if no improvement can be made.
\item Repeat from step 2.
\end{enumerate}

Rules are only generated if they can correctly lemmatize at least two examples in the training set. A dictionary is created for words which are incorrectly lemmatized by the rules, for example because they require a unique transformation, such as from \textit{við} ``we'' to \textit{ég} ``I''. Once trained, Nefnir lemmatizes words using the dictionary if they are present, or else with the most specific applicable rule.

A rule is generated for every suffix in a word form, with some restrictions. For base words, Nefnir considers all suffixes, from the empty string to the full word. For \textit{skó} ``shoes'', an inflected form of the word \textit{skór} ``shoe'', rules are generated for the suffixes $\varepsilon$, \textit{ó}, \textit{kó} and \textit{skó}. However, Nefnir does not create rules for suffixes that are shorter than the transformation required to lemmatize the word. For example, for \textit{bækur} ``books'', which requires the transformation \textit{ækur}$\to$\textit{ók} (the lemma for \emph{bækur} is \emph{bók}), only the suffixes \textit{ækur} and \textit{bækur} are considered.

Compounding is highly productive in Icelandic and compound words comprise a very large portion of the vocabulary. This is reflected in the DMII, where over 88\% of all words are compounds \cite{Bjarnadottir_2017}. Any of the open word classes can be combined to form a compound, and there is no theoretical limit to how many words they can consist of. Due to the abundance of compounds in the training data, and the freedom with which they can be formed, Nefnir places additional restrictions on which suffixes to consider when generating rules for them.
Suffixes for the final part of a compound are generated in the same manner as for base words, growing part by part thereafter. For example, the compound word \textit{fjall+göngu+skó} ``hiking boots'' would yield rules for the suffixes $\varepsilon$, \textit{ó}, \textit{kó}, \textit{skó}, \textit{gönguskó} and \textit{fjallgönguskó}. Allowing suffixes to grow freely past the final part of the compound may result in overfitting as the rules adapt to incidental patterns in the training data.

\section{Evaluation} \label{Evaluation}
%This will be based on the work already carried out by Svanhvít.
We have evaluated the output of Nefnir against a reference corpus of 21,093 tokens and their correct lemmas.

Samples for the reference corpus were extracted from two larger corpora, in order to obtain a diverse vocabulary:

\begin{itemize}
    \item The IFD corpus mostly contains literary texts \cite{Pind_1991}. It was first published in book form and is now available online. This corpus has been manually PoS tagged and lemmatized.
    \item The Icelandic Gold Standard (GOLD) is a PoS tagged and manually corrected corpus of around 1,000,000 tokens, containing a balanced sample of contemporary texts from 13 sources, including news texts, laws and adjucations, as well as various web content such as blog texts \cite{gullstadall_2010}. 
\end{itemize}

Samples were extracted at random from these two corpora, roughly 10,000 tokens from each, and the lemmas manually reviewed, following the criteria laid out in the preface of the IFD \cite{Pind_1991}.

The incentive when performing the evaluation was to create a diverse corpus of text samples containing foreign words, misspellings and other OOV words. Such words are likely to appear in real-world NLP tasks, and pose special problems for lemmatizers. In the proofread and literature-heavy IFD corpus, which was used for training and evaluating the previous two lemmatizers, these OOV words are less prevalent. Consequently, the test corpus used here is not directly comparable with the corpus used to evaluate Lemmald and the CST Lemmatizer for Icelandic. On the other hand, it is more diverse and offers more challenging problems for the lemmatizer.

One of the motivations of this work was to determine how well Nefnir performs when lemmatizing text which has been PoS tagged automatically, without any manual review, as such manual labour is usually not feasible in large-scale NLP tasks. For this purpose, we created two versions of the test corpus, one with the correct PoS tags, and another tagged using IceTagger \cite{Loftsson_2008}. The accuracy of IceTagger is further enhanced using data from the DMII. Measured against the correct PoS tags, the accuracy of the PoS tags in the reference corpus is 95.47\%.

Accuracy of the lemmatizaton was measured by comparing the reference corpus lemmas with the obtained lemmas from Nefnir. This was done for both the correctly tagged corpus (gold tags) and the automatically tagged one (IceTagger tags).
%Þarf eitthvað að útskýra betur hvað accuracy er hér?
As seen in Table \ref{table:2}, the accuracy for the test file with the correct PoS tags is 99.55\%, with 94 errors in 21,093 tokens. For the text tagged automatically with IceTagger, the accuracy is 96.88\%, with 658 errors. 

\begin{table}[]
\begin{tabular}{cc|cc}
\hline
\multicolumn{2}{c|}{\textbf{Gold tags}} & \multicolumn{2}{c}{\textbf{IceTagger tags}} \\ 
Accuracy (\%)     & Errors     & Accuracy (\%)       & Errors       \\ \hline
99.55        & 94              & 96.88           & 658\\\hline        
\end{tabular}
\caption{Results of the evaluation, with the accuracy and the total number of errors found.}
\label{table:2}
\end{table}

These results indicate that given correct PoS tags, Nefnir obtains high accuracy, with under a hundred errors in the whole corpus sample. This is comparable to the score reported for Lemmald, when DMII lookup has been added (99.55\%). In fact, it can be argued that a higher score is hard to come by, as natural language always contains some unforeseen issues that are hard to accommodate for, such as OOV words, misspellings, colloquialisms, etc. When Nefnir bases its lemmas on the automatically PoS tagged text, the accuracy decreases, from 99.55\% to 96.88\%, resulting in six times as many errors. 

We can classify the errors made by Nefnir into the following main categories:
\begin{enumerate}
    \item Foreign words
    \item Proper names
    \item Two valid lemmas for word form
    \item Typos
    \item Incorrect capitalization, abbreviations, hyphenation, etc.
    \item Unknown Icelandic words
    \item Wrong PoS tag leads to wrong lemma
\end{enumerate}

The most prevalent error categories when the PoS tags are correct are foreign words and proper names, such as foreign names of people, products and companies. A special issue that often came up is the cliticized definite article in Icelandic proper names. This is quite common in organization names (\textit{Síminn}, \textit{Samfylkingin}), titles of works of art (\textit{Svanurinn}), names of ships (\textit{Vonin}), buildings (\textit{Kringlan}), etc. Ultimately, it depends on the aim of the lemmatization how these should be handled, but in this evaluation we assume as a general rule that they should be lemmatized with the definite article (\textit{Síminn}, and not \textit{sími} or \textit{Sími}). The same applies to the plural, in names such as \textit{Hjálmar} ``helmets'' (band) and \textit{Katlar} (place name).

In the automatically tagged data, tagging errors are the most common source of lemmatization errors, such as when \textit{læknum} (referring to the plural dative of the masculine noun \textit{læknir} ``doctor'') is tagged as being in the singular, which leads to it being incorrectly lemmatized as \textit{lækur} ``brook''. This was to be expected, as the rules learned from the DMII rely on the correct tagging of the input. However, as the authors of Lemmald comment, as long as the word class is correct, the lemmatizer can usually still find the correct lemma \cite{Ingason_2008}.

The main reason for the high accuracy in our view lies in the richness of the DMII data. No lexicon can ever include all words of a particular language, as new words appear every day, but most often, new words in Icelandic are compounds, created from words already present in the DMII. This explains how rare or unknown words such as the adjective \textit{fuglglaður} ``bird-happy'', which appears in the corpus data, can be correctly lemmatized using the suffix rule for \textit{glaður} ``happy''.

As mentioned above, Nefnir, the CST Lemmatizer for Icelandic, and Lemmald have not been evaluated using the same reference corpus. The accuracy of the three lemmatizers are, therefore, not directly comparable, but our results indicate that Nefnir obtains the highest accuracy.%, especially given the diversity of the test corpus.

\section{Conclusion}
We described and evaluated Nefnir, a new open source lemmatizer for Icelandic. It uses suffix substitution rules, derived from a large morphological database, to lemmatize tagged text.  Evaluation shows that Nefnir obtains high accuracy for both correctly and automatically PoS-tagged input.

As taggers for Icelandic gradually get better, we can expect to see the lemmatization accuracy go up as well. Expanding the morphological database with more proper names may also help to achieve even higher accuracy.
%Hvað með KVIST?!

%\section{Translation of non-English Terms}

%It is also advised to supplement non-English characters and terms
%with appropriate transliterations and/or translations
%since not all readers understand all such characters and terms.
%Inline transliteration or translation can be represented in
%he order of: original-form transliteration ``translation''.

%\section*{Acknowledgments}

%Do not number the acknowledgment section. Do not include this section
%when submitting your paper for review.

\bibliographystyle{acl_natbib}
\bibliography{nefnir}

\end{document}